\documentclass[hasAbstract,authorBox]{csmagazine}

\usepackage{times}
\usepackage{latexsym}
\usepackage{MnSymbol}
\usepackage{amsmath}
\usepackage{amsfonts}
\usepackage{url}
\usepackage{color,soul}
\usepackage{flushend}
\usepackage{graphicx}
\usepackage{multirow}
\usepackage{microtype}

\usepackage{hyperref}
\usepackage[capitalize]{cleveref}

\DeclareUnicodeCharacter{1EBF}{\'{\^{e}}}

\title{Sentiment and Sarcasm Classification with Multitask Learning}

\author{Navonil~Majumder\\ \ieeecsAffiliation{Instituto Polit\'ecnico Nacional}\\
	Soujanya~Poria\\ \ieeecsAffiliation{Nanyang Technological University}\\
	Haiyun~Peng\\ \ieeecsAffiliation{Nanyang Technological University}\\ 
	Niyati~Chhaya\\ \ieeecsAffiliation{Adobe Research}\\  
	Erik~Cambria\\ \ieeecsAffiliation{Nanyang Technological University}\\  
	Alexander~Gelbukh\\ \ieeecsAffiliation{Instituto Polit\'ecnico Nacional}\\ 
	Editor: \\Erik Cambria\\\ieeecsAffiliation{Nanyang Technological University}
}

\begin{document}

\ieeecsPageHeaders{AFFECTIVE COMPUTING AND SENTIMENT ANALYSIS}{IEEE INTELLIGENT SYSTEMS}{DEPARTMENT}

\ieeecsArticleTitle

\ieeecsInsertAuthor

\raggedright

\ieeecsAbstract{Sentiment classification and sarcasm detection are both important natural language processing (NLP) tasks. Sentiment is always coupled with sarcasm where intensive emotion is expressed. Nevertheless, most literature considers them as two separate tasks. We argue that knowledge in sarcasm detection can also be beneficial to sentiment classification and vice versa. We show that these two tasks are correlated, and present a multi-task learning-based framework using a deep neural network that models this correlation to improve the performance of both tasks in a multi-task learning setting. Our method outperforms the state of the art by 3--4\% in the benchmark dataset.}

%Computer Society style is \RaggedRight for the article body
\raggedright

The surge of Internet has enabled large-scale text-based opinion sharing on a
wide range of topics. This has led to the opportunity of mining user sentiment
on various subjects from the data publicly available over the Internet. The most
important task in the analysis of users' opinions is sentiment
classification: determining whether a given text, such as a user review, comment,
or tweet, carries positive or negative polarity. 

When expressing their opinions, users often use sarcasm for
emphasizing their sentiment. In a sarcastic text, the sentiment intended by the author is the
opposite of its literal meaning. For example, the sentence ``\emph{Thank you alarm for never going off}'' is literally positive
% the sentence starts with positive sentiment 
(``\emph{Thank you}''),
however, the intended sentiment is negative ``\emph{alarm never going off}.''
% , which entails overall negative sentiment. 
Unless this sentiment shift is detected with 
% the surface 
semantics, the classifier may fail to spot sarcasm.
%This means, detecting
%sarcasm is another important task required for correct text understanding. 

Currently, most researchers focus on either sentiment classification or sarcasm
detection,\cite{porloo,joshi2015harnessing} without considering the possibility
of mutual influence between the two tasks. However, one can observe that the two
tasks are correlated: people often use sarcasm as a device
for the expression of emphatic negative sentiment. This observation can lead to a
simple way in which one of the two tasks can help improve the other, i.e., if an
expression can be detected as sarcastic, its sentiment can be assumed negative;
if the expression can be classified as positive, then it can be assumed not
sarcastic.
\raggedright

Here, we show that while this logic does lead to a slight improvement, there is
a better way of combining the two tasks. Namely, in this paper, we train a
classifier for both sarcasm and sentiment in a single neural network using
multi-task learning, a novel learning scheme that has gained recent
popularity\cite{augenstein-sogaard:2017:Short,lan-EtAl:2017:EMNLP20172}.
We empirically show that this method outperforms the results obtained with two
separate classifiers and, in particular, outperforms the current state of the
art by Mishra et al.\cite{mishra-dey-bhattacharyya:2017:Long}

The remainder of the paper is organized as follows: \cref{sec:related} outlines related work; \cref{sec:method} presents our approach; \cref{sec:experiments} lists the baselines; \cref{sec:results-discussions} discusses results; finally, \cref{sec:conclusions} concludes the paper.

\section{Related Work}\label{sec:related}

% In sentiment analysis, it generally falls into two categories: statistical
% methods and knowledge-based approaches\cite{cambria2016affective}.
%Recent statistical methods mainly rely on word
% embeddings\cite{mikolov2013distributed} and deep learning techniques.

Machine learning methods and deep neural networks, such as convolutional, recursive, recurrent, and memory networks, have shown good performance for sentiment detection.\cite{zadeh2018multi,majumder2019dialoguernn,dong2014adaptive,sutskever2014sequence} Knowledge-based methods explore syntactic patterns\cite{poria2015sentiment} and employ sentiment resources.\cite{camnt5} However, sarcasm detection currently focuses on extracting
features, such as syntactic,\cite{barbieri2014modelling} 
surface pattern-based,\cite{davidov2010semi}
or personality-based features,\cite{porloo} as well as
contextual incongruity.\cite{joshi2015harnessing}

Mishra et al.\cite{mishra-dey-bhattacharyya:2017:Long} extracted multimodal cognitive
features for both sentiment classification and sarcasm detection, without modeling the two tasks in a single system. However, recently multi-task learning has been successfully applied
in many NLP tasks, such as implicit discourse relationship 
identification\cite{lan-EtAl:2017:EMNLP20172} and key-phrase boundary
classification.\cite{augenstein-sogaard:2017:Short} In this paper, we apply it to sentiment classification and sarcasm detection.
% This substantiates our
% assumption of the effectiveness of multi-task learning based
% architecture for both sarcasm detection and sentiment classification.

\section{Method}
\label{sec:method}
According to Riloff et al.,\cite{Riloff2013SarcasmAC} most sarcastic sentences carry 
negative sentiment. We leverage this to improve both sentiment classification and sarcasm detection.
%So, we hypothesize that sentimental information will
%contribute to better sarcasm detection, since positive sentences are less likely
%to be sarcastic.
We use multi-task learning, where a single neural
network is used to perform more than one classification task (in our case,
sentiment classification and sarcasm detection). This network facilitates synergy
between the two tasks, resulting in improved performance on both
tasks in comparison with their standalone counterparts.

% In the following sections, we discuss our architecture and the motivations in details.

\begin{figure}[ht]
	\centering
	\includegraphics[width=0.48\textwidth]{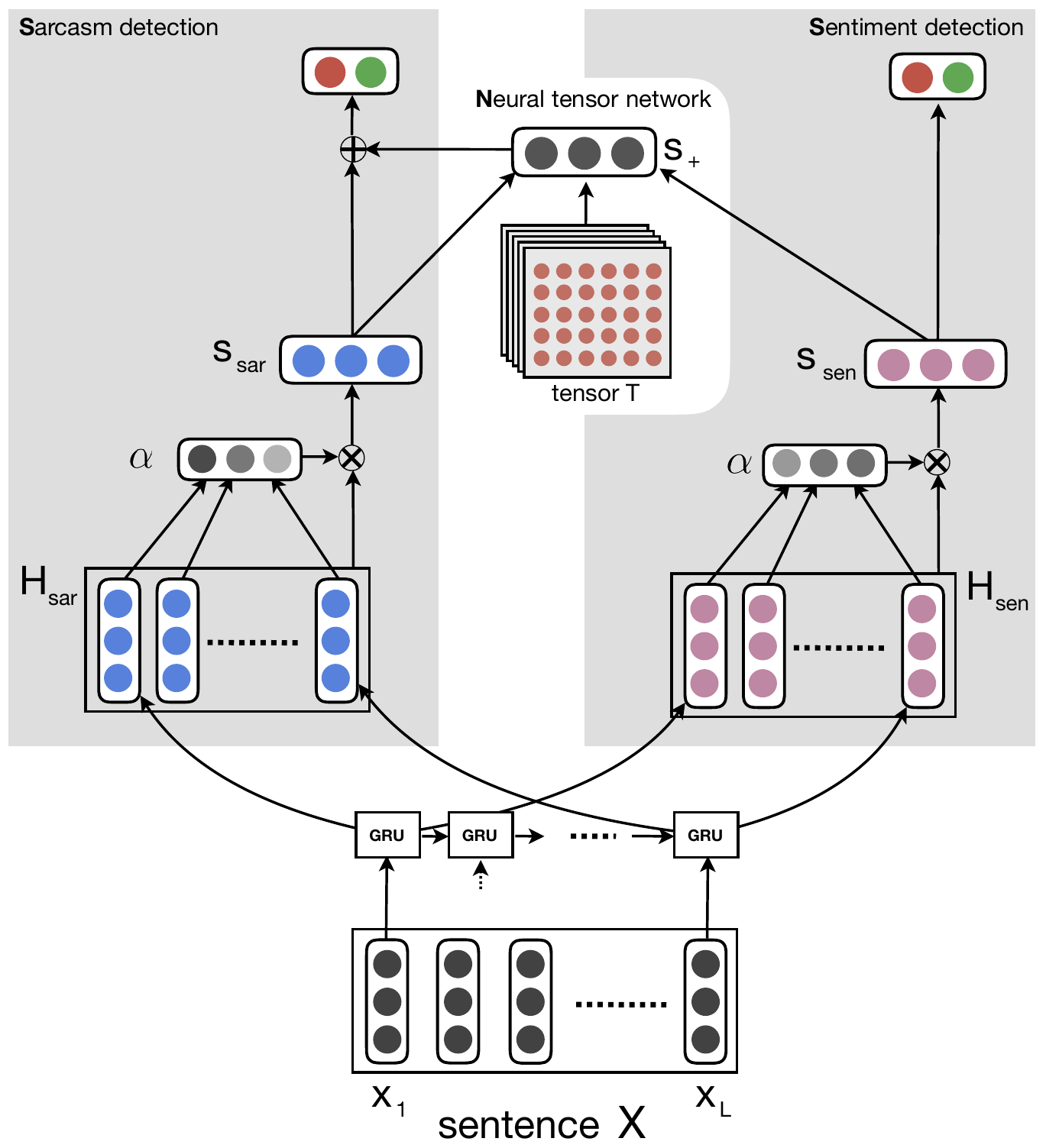}
	\caption{Our multi-task architecture.}
	\label{fig:architecture}
\end{figure}

%Our method can be summarized as follows.
%We evaluated our architecture on the same dataset as was used by\cite{mishra-dey-bhattacharyya:2017:Long}, which
%lists sentences annotated with both sentiment and sarcasm labels. 
%We first represent the words in the sentences with Glove word embeddings\cite{pennington2014glove} and then feed them fed to a Gated Recurrent Unit (GRU) to obtain context-aware word
%representations; see \cref{fig:architecture}. 
%Next, we transform these word representations to sentiment- and
%sarcasm-specific representations using two fully-connected layers. 
%We obtain task-specific sentence representations by applying attention network on the word
%representations.
%For sentiment classification, we feed the sentiment-specific sentence representation to
%a softmax classifier. 
%For sarcasm detection, we first apply a Neural Tensor Network (NTN) on both task-specific sentence representations for inter-task communication, and obtain a joint sentence representation.
%Finally, we concatenate this joint representation with sarcasm representation for sarcasm classification with a softmax layer.

\paragraph{Task Definition}
%\label{sec:problem-definition}

We solve two tasks with a single network. Given a sentence
%\hl{for matrices, () or [] ?}
$[w_1,w_2,\dots,w_l]$, where $w_i$ are words, we assign it both a sentiment tag
(positive / negative) and a sarcasm tag (yes / no).

\paragraph{Input Representation}
%\label{sec:word-embeddings}

We use $D_g$-dimensional ($D_g=300$) Glove
word-embeddings $x_i\in\mathbb{R}^{D_g}$ to represent
the words $w_i$,
padding the variable-length input sentences to a fixed length with null vectors. 
Thus, the input is
represented as a matrix $X= [x_1,x_2,\dots,x_L]$, where $L$ is the length of
% padded sentences, which is the length of 
the longest sentence.
% , and $x_i\in\mathbb{R}^{D_g}$ are Glove word-embeddings for $i=1,2,\dots,L$.

\paragraph{Sentence Representation}
% \label{sec:sent-repr}
In the next layers, we obtain sentence representation from $X$ using gated recurrent unit (GRU) with attention mechanism as explained below.

\paragraph{\it Sentence-level word representation}
% \label{sec:sentence-level-word}
%\hl{shorten this by citing another paper}
The sentence $X$ is fed to a GRU of size $D_{gru}=500$ with parameters 
$W^{[z, r, h]}\in \mathbb{R}^{D_g\times D_{gru}}$ and $U^{[z, r, h]}\in \mathbb{R}^{D_{gru}\times
	D_{gru}}$
% and $z_t, r_t, h_t, s_t \in \mathbb{R}^{D_{gru}}$ 
%to get context-rich
%sentence-level word representations $H= [h_1,h_2,\cdots,h_L]$ ($H$ is the hidden
%output of the GRU).
% \begin{flalign*}
% z&=\sigma(x_{t}U^{z}+s_{t-1}W^{z}),\\
% r&=\sigma(x_{t}U^{r}+s_{t-1}W^{r}),\\
% h_{t}&=\tanh(x_{t}U^{h}+(s_{t-1}*r)W^{h}),\\
% s_{t}&=(1-z)*h_{t}+z*s_{t-1},
% \end{flalign*}
% where $U^{[z, r, h]}\in \mathbb{R}^{D_g\times D_{gru}}$, $W^{[z, r, h]}\in \mathbb{R}^{D_{gru}\times
% D_{gru}}$, and $z, r, h_t, s_t \in \mathbb{R}^{D_{gru}}$.
to get context-rich
sentence-level word representations $H= [h_1,h_2,\cdots,h_L]$,
$h_t \in \mathbb{R}^{D_{gru}}$ at the hidden
output of the GRU.

We use $H$ for both sarcasm and sentiment. 
Thus, $H$ is transformed to $H_{sar}$ and $H_{sen}$ using two different
fully-connected layers of size $D_t=300$ in order to accommodate two
different tasks, sarcasm detection and sentiment classification:
\begin{flalign*}
	H_{sar}&=ReLU(H W_{sar}+b_{sar}),\\
	H_{sen}&=ReLU(H W_{sen}+b_{sen}),
\end{flalign*}
where $W_{[sar,sen]}\in \mathbb{R}^{D_{dru}\times D_{t}}$ and $b_{[sar,sen]}\in
\mathbb{R}^{D_{t}}$.

\paragraph{\it Attention network}
%\label{sec:attention-network}

Word representations in $H_*$ are encoded with task-specific
sentence-level context. To aggregate these context-rich representations into the sentence representation $s_*$, we use attention mechanism, due to its ability to prioritize
words relevant for the classification:
\begin{flalign}
	P&=\tanh(H_* W^{ATT}),\\
	\alpha&=softmax(P^T W^\alpha), \label{eqn:attn_alpha}\\
	s_*&=\alpha H_*^T, \label{eqn:sent_rep}
\end{flalign}
where $W^{ATT}\in \mathbb{R}^{D_t\times 1}$, $W^\alpha\in \mathbb{R}^{L\times
	L}$, $P\in \mathbb{R}^{L\times 1}$, and $s_*\in
\mathbb{R}^{D_t}$.
In \cref{eqn:attn_alpha}, $\alpha \in [0,1]^L$ gives the relevance
of words for the task, multiplied 
in \cref{eqn:sent_rep} by the context-aware word representations in $H_*$.
% by the corresponding relevance in $\alpha$.
% and sum them to obtain sentence representation~$s_*$.

\paragraph{Inter-Task Communication}
%\label{sec:information-exchange}

We use neural tensor network (NTN) of size $D_{ntn}=100$ to
fuse sarcasm- and sentiment-specific sentence representations, $s_{sar}$
and $s_{sen}$, to obtain the fused representation $s_+$, where
\begin{flalign*}
	\!\!
	s_+=\tanh(s_{sar} T^{[1:D_{ntn}]} s_{sen}^T+(s_{sar}\oplus s_{sen}) W+b),
\end{flalign*}
where $T\in \mathbb{R}^{D_{ntn}\times D_t\times D_t}$, $W\in \mathbb{R}^{2D_t\times
	D_{ntn}}$, $b, s_+ \in \mathbb{R}^{D_{ntn}}$, and
$\oplus$ stands for concatenation. The vector $s_+$ contains information
relevant to both sentiment and sarcasm. Instead of NTN, we also tried
attention and concatenation for fusion, which resulted in inferior performance
(\cref{sec:results-discussions}).

\paragraph{Classification}
%\label{sec:classification}

For the two tasks, we use two different softmax
layers for classifications.

\paragraph{\it Sentiment classification}
We use only $s_{sen}$ as sentence representation for sentiment
classification, since we observe best performance without $s_+$. We apply
softmax layer of size $C$ ($C=2$ for binary task) on $s_{sen}$ for classification as follows:
\begin{flalign*}
	\mathcal{P}_{sen}&=\text{softmax}(s_{sen}~W_{sen}^{softmax}+b_{sen}^{softmax}),\\
	\hat{y}_{sen}&=\underset{j}{\text{argmax}}(\mathcal{P}_{sen}[j]),
\end{flalign*}
where $W_{sen}^{softmax}\in \mathbb{R}^{D_t\times C}$, $b_{sen}^{softmax}\in
\mathbb{R}^C$, $\mathcal{P}_{sen}\in \mathbb{R}^C$, $j$ is the class value
(0 for negative and 1 for positive), and $\hat{y}_{sen}$ is the estimated class value.

\paragraph{\it Sarcasm classification}
We use $s_{sar}\oplus s_+$ as sentence representation for sarcasm
classification using softmax layer with size $C$ ($C=2$) as follows:
\begin{flalign*}
	\!\!\mathcal{P}_{sar}&=\text{softmax}((s_{sar}\oplus s_+)~W_{sar}^{softmax}+b_{sar}^{softmax}),\\
	\!\!\hat{y}_{sar}&=\underset{j}{\text{argmax}}(\mathcal{P}_{sar}[j]),
\end{flalign*}
where $W_{sar}^{softmax}\in \mathbb{R}^{(D_t+D_{ntn})\times C}$, $b_{sar}^{softmax}\in
\mathbb{R}^C$, $\mathcal{P}_{sar}\in \mathbb{R}^C$, $j$ is the class value
(0 for no and 1 for yes), and $\hat{y}_{sar}$ is the estimated class value.

\paragraph{Training}
%\label{sec:training}

We use categorical cross-entropy as the loss function ($J_*$; * is $sar$ or $sen$) for training:
\begin{flalign*}
	J_*=-\dfrac{1}{N}\sum_{i=1}^N\sum_{j=0}^{C-1}{y^*_{ij}~\log \mathcal{P}_{*i}[j]},
\end{flalign*}
where $N$ is the number of samples, $i$ is the index of a sample, $j$ is the
class value, and
\begin{flalign*}
	y^*_{ij}=
	\begin{cases}
		1, \text{ if expected class value of sample }i\text{ is }j,\\
		0, \text{ otherwise}.
	\end{cases}
\end{flalign*}

For training, we use ADAM,\cite{DBLP:journals/corr/KingmaB14} an algorithm based on stochastic gradient descent which optimizes each
parameter individually with different and adaptive learning rates. Also, we
minimize both loss functions, namely $J_{sen}$ and $J_{sar}$, with equal
priority, by optimizing the parameter set
\begin{flalign*}
	\theta=&\{U^{[z,r,h]},W^{[z,r,h]},W_*,b_*,W^{ATT},\\
	&W^\alpha,T,W,b,W^{softmax}_*,b^{softmax}_*\}.
\end{flalign*}

%\vspace*{-0.8cm}
\section{Experiments}
\label{sec:experiments}

\paragraph{Dataset}
%\label{sec:dataset-details}

The dataset\cite{AAAI1612070} consists of 994 samples, each sample containing a text snippet
labeled with sarcasm tag, sentiment tag, and eye-movement data of 7 readers. We
ignored the eye-movement data in our experiments. Of those samples, 383 are positive
and 350 are sarcastic.

\paragraph{Baselines and Model Variants}
\label{sec:baselines}

We evaluated the following baselines and variations of our model.

% In the following paragraphs, we describe them in detail.

\paragraph{\it Standalone classifiers}

Here, we used
\begin{flalign*}
	h_{*}=&FCLayer(GRU(X)),\\
	\mathcal{P}=&SoftmaxLayer(h_{*}),
\end{flalign*}
where * represents $sar$ or $sen$,
$X$ is the input sentence as a list of word embeddings. We feed $X$ to GRU
and pass the final output through a fully-connected layer ($FCLayer$) to obtain
sentence representation $h_{*}$. We apply final softmax classification
($SoftmaxLayer$) to $h_{*}$.

\paragraph{\it Sentiment coerced by sarcasm}

In this classifier, the sentences classified as sarcastic are forced to be
considered negative by the sentiment classifier.

%\paragraph{Sentiment coerced by sarcasm}
%ADD VICE VERSA: positive are forced to be not sarcastic

\paragraph{\it Simple multi-task classifier} %13

The following equations 
% roughly ??????
summarize this variant:
\begin{flalign}
	h_*=&FCLayer_*(GRU(X)),            \label{eq:H_*=}\\
	\mathcal{P}_*=&SoftmaxLayer_*(h_*),\label{eq:P_*=}
\end{flalign}
where * represents $sar$ or $sen$. This setting shares the GRU between two
tasks. Final output of GRU is taken as the sentence representation. Sentence
representation is fed to two different task-specific fully-connected layers
($FCLayer_*$), giving $h_{*}$. Subsequently, $h_{*}$ are fed to two different
softmax layers $SoftmaxLayer_*$ for classification.

% \paragraph{Shared Fully-Connected Layer based Fusion.} %14

% The setting can be summarized as follows:
% {\small\begin{flalign*}
%     H_*=&FCLayer_*(GRU(X)),\\
%     H2_*=&FCLayer2(H_*),\\
%     \mathcal{P}_*=&SoftmaxLayer_*(H2_*),
% \end{flalign*}
% }where * represents $sar$ or $sen$.
% This setting is same as the one discussed previously, except for $H_{sar}$ and
% $H_{sem}$ are fed to one additional shared fully-connected layer $FCLayer2$,
% whose outputs are used for classification.

\paragraph{\it Simple multi-task classifier with fusion} %1

In this variant, we changed~\cref{eq:P_*=} to:
\begin{flalign}
	\mathcal{P}_{sar}=&SoftmaxLayer_{sar}(h_{sar}\oplus F),\label{eq:P_sar}\\
	\mathcal{P}_{sen}=&SoftmaxLayer_{sen}(h_{sen}),\label{eq:P_sen}
\end{flalign}
where $F=NTN(h_{sar},h_{sen})$. Here, $h_{sar}$ and $h_{sem}$ are
fed to a NTN whose output is concatenated with
$h_{sar}$ for classification. Sentiment classification is done with $h_{sen}$
only.
We also tried variants with
other methods of fusion (such as fully connected layer or Hadamard product) instead of NTN, as well as variants
with $h_{sen}\oplus F$ instead of, or in addition to, 
$h_{sar}\oplus F$,
%\begin{flalign*}
%\mathcal{P}_{sar}=&SoftmaxLayer_{sar}(H_{sar}),\\
%\mathcal{P}_{sen}=&SoftmaxLayer_{sen}(H_{sen}\oplus F)
%\end{flalign*}
%and
%\begin{flalign*}
%\mathcal{P}_{sar}=&SoftmaxLayer_{sar}(H_{sar}\oplus F),\\
%\mathcal{P}_{sen}=&SoftmaxLayer_{sen}(H_{sen}\oplus F),
%\end{flalign*}
but they did not imprive the results.
%, as 
% for the reasons 
% discussed in \cref{sec:results-discussions}.

\paragraph{\it Task-specific GRU with fusion} %15

Here, we used two separate GRUs for the two tasks in~\cref{eq:H_*=}:
\begin{flalign}
	h_*=&FCLayer_*(GRU_*(X)).\label{eq:GRU_*}
\end{flalign}
We used \cref{eq:P_sar} and \cref{eq:P_sen} for $\mathcal{P}_*$. Again, we tried concatenating
$F$ with $h_{sen}$, both, or none as in \cref{eq:P_*=}, but this did not improve the results.
% Again, we also tried concatenating $F$ with $H_{sar}$, which did not improve the results.

\begin{table*}[t]
	\centering
	%\small
	\tiny
	\caption{Results for various experiments.}
	\label{tab:results}
	\resizebox{\textwidth}{!}{
		\begin{tabular}{lccccccc}
			\hline
			\multirow{2}{*}{Variant} & \multicolumn{3}{c}{Sentiment} & \multicolumn{3}{c}{Sarcasm} & Average\\
			\cline{2-8} & Precision & Recall & F-Score & Precision & Recall & F-Score & F-Score \\
			\hline
			State of the art\cite{mishra-dey-bhattacharyya:2017:Long} & 79.89 & 74.86 & 77.30 & 87.42 & 87.03 & 86.97 & 82.13
			\\
			\hline
			Standalone classifiers & 79.02 & 78.03 & 78.13 & 89.96 & 89.25 & 89.37 & 83.75
			\\
			Standalone coerced & 81.57 & 80.06 & 80.38 & -- & -- & -- & -- \\
			\hline
			Multi-Task simple & 80.41 & 79.88 & 79.7 & 89.42 & 89.19 & 89.04 & 84.37 \\
			Multi-Task with fusion & 82.32 & 81.71 & 81.53 & 90.94 & \textbf{90.74} & \textbf{90.67} & 86.10
			\\
			% Shared FC-Based Fusion & 80.54 & 79.8 & 79.73 & 89.28 & 89.05 & 88.91 \\
			% Cross-Attention & 82.89 & 82.53 & 82.37 & 90.57 & 90.31 & 90.27 \\
			% Non-Shared Attention & 82.93 & 82.39 & 82.31 & 90.33 & 90.02 & 90.01 \\
			Multi-Task with fusion and separate GRUs & 80.54 & 80.02 & 79.86 & \textbf{91.01} & 90.66 & 90.62 & 85.24
			\\
			Multi-Task with fusion and shared attention (\cref{sec:method}) & \textbf{83.67} & \textbf{83.10} & \textbf{83.03} & 90.50 & 90.34 & 90.29 & \bf 86.66
			\\
			% Word-Level NTN & \textbf{83.12} & \textbf{82.56} & \textbf{82.48} & \textbf{82.56} & 90.93 & 90.69 & 90.66 & 90.69\\
			\hline
		\end{tabular}}
	\end{table*}

	\paragraph{\it Best model: shared attention} %12
	
	% This setting can be summarized as follows:
	% {\small\begin{flalign*}
	% H_*=&FCLayer_*(GRU(X)),\\
	% R_*=&Attention(H_*),\\
	% F=&NTN(R_{sar},R_{sen}),\\
	% \mathcal{P}_{sar}=&SoftmaxLayer_{sar}(R_{sar}\oplus F),\\
	% \mathcal{P}_{sen}=&SoftmaxLayer_{sen}(R_{sen}),
	% \end{flalign*}
	% }where * represents $sar$ or $sen$. We use attention ($Attention$) to obtain the
	%   sentence representation.
	Here, we added the attention mechanism to the matrix $H_*$ in~\cref{eq:H_*=}, and used~\cref{eq:P_sar} and~\cref{eq:P_sen} for $\mathcal P_*$.
	This model, described in detail in \cref{sec:method}, is the main model we present in this paper since it gave the best results. We also tried separate GRUs as in~\cref{eq:GRU_*}, but this did not improve the results.

	\section{Results and Discussion}
	\label{sec:results-discussions}
	
	The results using 10-fold cross validation are shown in \cref{tab:results}. 
	As baselines, we used the standalone sentiment and sarcasm classifiers, as well as the CNN-based 
	state-of-the-art method by Mishra et al.\cite{mishra-dey-bhattacharyya:2017:Long} Our standalone GRU-based sentiment and sarcasm classifiers performed slightly better than the state of the art, even though this also uses the gaze data present in the dataset
	% , which we ignored in our experiments, yet our classifiers have performed better.
	but this is hardly available in any real-life setting. 
	In contrast, our method, besides improving results, is applied to plain-text documents such as tweets, without any gaze data.
	
	As expected, the sentiment classifier coerced by sarcasm classifier
	performed better than the standalone sentiment classifier. This means that an
	efficient sarcasm detector can boost the performance of a sentiment
	classifier. 
	All our multi-task classifiers outperformed both standalone classifiers.
	However, the margin of improvement for multi-task classifier over the
	standalone classifier is greater for sentiment than for sarcasm.
	Probably this is because sarcasm detection is a subtask of sentiment analysis.\cite{camsui}
	
	Analyzing examples and attention visualization of the multi-task network, we observed that the multi-task network mainly helps improving sarcasm classification when there is a strong sentiment shift, which indicates the possibility of sarcasm in the sentence. 
	The example given in the introduction was classified incorrectly by the
	standalone sarcasm classifier but correctly by the standalone sentiment classifier; coercing one of the classifiers by the other would not change the result.
	In the multi-task network, both sentiment and sarcasm are detected correctly,
	apparently because 
	% as attention visualization (not presented in this short paper) shows, 
	the network detected the sentiment shift in the sentence, which improved sarcasm classification.
	
	Similarly, the sentence ``\emph{Absolutely love when water is spilt on my phone, just love it}'' is classified as positive by the standalone sentiment classifier: ``\emph{Absolutely love}'' highlighted by the attention scores (not presented in this short paper). 
	However, the standalone sarcasm classifier
	identified it as sarcastic due to ``\emph{water spilt on my phone}'' (seen
	from the attention scores) and in the multi-task network this clue corrected the
	sentiment classifier's output.
	
	%This further illustrates the claim of\cite{porloo} that
	%sentiment can be determined with sarcasm.
	% sarcasm -> sentiment cambria et al.
	%sentiment shift apparent poria et al.
	
	Even our standalone GRU-based classifiers outperformed the CNN-based state-of-the-art method.
	The multi-task classifiers outperformed the standalone classifiers 
	% We believe that this was because the shared GRU learns representations for both tasks.
	% -- NO: separate GRUs also outperform separate
	because of the shared representation, which serves as additional regularization for each task from the other task.
	% The last example was classified correctly by this setup, which is a testament of effectiveness of multi-task setup. -- SAID ALREADY
	% THIS WAS SAID ABOUT THE OTHER EXAMPLE:
	% We assume that this network was able to identify the sudden shift of
	% sentiment from ``\emph{Absolutely love}'' to ``\emph{water is spilt on my
	% phone}'', which is a strong sign of sarcasm, as claimed by\cite{suitcase}.
	
	Adding NTN fusion to the multi-task classifier further improved
	results, giving the best performance for sarcasm detection. Adding an attention network shared between the tasks
	further improves the performance for sentiment classification.
	As the last column of \cref{tab:results} shows, on average the best results across the two tasks were obtained with the architecture described in~\cref{sec:method}.
	
	% \hl{mention why sentiment improves more than sarcasm}
	
	\section{Conclusion}
	\label{sec:conclusions}
	
	We presented a classifier architecture that can be trained on sentiment or sarcasm data and
	outperforms the state of the art in both cases on the dataset used by Mishra et al.\cite{mishra-dey-bhattacharyya:2017:Long} Our architecture uses a GRU-based neural network, while the state-of-the-art method used a CNN.
	
	Furthermore, we showed that multi-task learning-based methods significantly
	outperform standalone sentiment and sarcasm classifiers. 
	This indicates that sentiment classification and sarcasm detection are related tasks.
	
	Finally, we presented a multi-task learning architecture that gave the best results, out of a number of variants of the architecture that we tried.
	
	To make our claim more robust, we plan to build a new dataset for rigorous experimentation. 
	In addition, we intend to incorporate multimodal information in our network for enhancing its performance.

%%%%%%%%%%%%%%%%%%%%%%%%%%%%%%%%%%%%%%%%%%%%%%%%%%%%%%%%%%%%%
%%                  Acknowledgments                        %%
%%                                                         %%
%%                                                         %%
%%%%%%%%%%%%%%%%%%%%%%%%%%%%%%%%%%%%%%%%%%%%%%%%%%%%%%%%%%%%%


\ieeecsReferences{REFERENCES}

%Usage of the ieeeCSBib reference style file is required
\bibliographystyle{ieeeCSBib}

%change refs to the name of your .bib file
\bibliography{sentiment-and-sarcasm-classification-with-multitask-learning}

\begin{thebibliography}{10}
\newcommand{\enquote}[1]{``#1''}
\providecommand{\url}[1]{\texttt{#1}}
\providecommand{\urlprefix}{}
\expandafter\ifx\csname urlstyle\endcsname\relax
  \providecommand{\doi}[1]{doi:\discretionary{}{}{}#1}\else
  \providecommand{\doi}{doi:\discretionary{}{}{}\begingroup
  \urlstyle{rm}\Url}\fi

\bibitem{porloo}
S.~Poria et~al., \enquote{A Deeper Look into Sarcastic Tweets Using Deep
  Convolutional Neural Networks,} \emph{{COLING}}, 2016, pp. 1601--1612.

\bibitem{joshi2015harnessing}
A.~Joshi, V.~Sharma, and P.~Bhattacharyya, \enquote{Harnessing Context
  Incongruity for Sarcasm Detection.} \emph{ACL}, 2015, pp. 757--762.

\bibitem{augenstein-sogaard:2017:Short}
I.~Augenstein and A.~S{\o}gaard, \enquote{{Multi-Task Learning of Keyphrase
  Boundary Classification},} \emph{{ACL}}, 2017, pp. 341--346.

\bibitem{lan-EtAl:2017:EMNLP20172}
M.~Lan et~al., \enquote{{Multi-task Attention-based Neural Networks for
  Implicit Discourse Relationship Representation and Identification},}
  \emph{EMNLP}, 2017, pp. 1299--1308.

\bibitem{mishra-dey-bhattacharyya:2017:Long}
A.~Mishra, K.~Dey, and P.~Bhattacharyya, \enquote{Learning Cognitive Features
  from Gaze Data for Sentiment and Sarcasm Classification using Convolutional
  Neural Network,} \emph{ACL}, 2017, pp. 377--387.

\bibitem{zadeh2018multi}
A.~Zadeh et~al., \enquote{Multi-attention recurrent network for human
  communication comprehension,} \emph{AAAI}, 2018, pp. 5642--5649.

\bibitem{majumder2019dialoguernn}
N.~Majumder et~al., \enquote{DialogueRNN: An Attentive RNN for Emotion
  Detection in Conversations,} \emph{AAAI}, 2019.

\bibitem{dong2014adaptive}
L.~Dong et~al., \enquote{{Adaptive Recursive Neural Network for
  Target-dependent Twitter Sentiment Classification.}} \emph{ACL}, 2014, pp.
  49--54.

\bibitem{sutskever2014sequence}
I.~Sutskever, O.~Vinyals, and Q.~V. Le, \enquote{Sequence to sequence learning
  with neural networks,} \emph{NIPS}, 2014, pp. 3104--3112.

\bibitem{poria2015sentiment}
S.~Poria et~al., \enquote{Sentiment data flow analysis by means of dynamic
  linguistic patterns,} \emph{IEEE Computational Intelligence Magazine},
  vol.~10, no.~4, 2015, pp. 26--36.

\bibitem{camnt5}
E.~Cambria et~al., \enquote{{SenticNet} 5: Discovering conceptual primitives
  for sentiment analysis by means of context embeddings,} \emph{{AAAI}}, 2018,
  pp. 1795--1802.

\bibitem{barbieri2014modelling}
F.~Barbieri, H.~Saggion, and F.~Ronzano, \enquote{Modelling Sarcasm in Twitter,
  a Novel Approach.} \emph{WASSA@ACL}, 2014, pp. 50--58.

\bibitem{davidov2010semi}
D.~Davidov, O.~Tsur, and A.~Rappoport, \enquote{Semi-supervised recognition of
  sarcastic sentences in Twitter and Amazon,} \emph{CoNLL}, 2010, pp. 107--116.

\bibitem{Riloff2013SarcasmAC}
E.~Riloff et~al., \enquote{{Sarcasm as Contrast between a Positive Sentiment
  and Negative Situation},} \emph{EMNLP}, 2013, pp. 704--714.

\bibitem{DBLP:journals/corr/KingmaB14}
D.~P. Kingma and J.~Ba, \enquote{Adam: {A} Method for Stochastic Optimization,}
  \emph{CoRR}, vol. abs/1412.6980, 2014.

\bibitem{AAAI1612070}
A.~Mishra, D.~Kanojia, and P.~Bhattacharyya, \enquote{Predicting Readers'
  Sarcasm Understandability by Modeling Gaze Behavior,} \emph{AAAI}, 2016, pp.
  3747--3753.

\bibitem{camsui}
E.~Cambria et~al., \enquote{Sentiment Analysis is a Big Suitcase,} \emph{{IEEE}
  Intelligent Systems}, vol.~32, no.~6, 2017, pp. 74--80.

\end{thebibliography}

%%%%%%%%%%%%%%%%%%%%%%%%%%%%%%%%%%%%%%%%%%%%%%%%%%%%%%%%%%%%%
%%                  Author Bios                            %%
%%                                                         %%
%%                                                         %%
%%%%%%%%%%%%%%%%%%%%%%%%%%%%%%%%%%%%%%%%%%%%%%%%%%%%%%%%%%%%%

\ieeecsAboutAuthor{ABOUT THE AUTHORS}

\begin{ieeecsAuthorBio}
\textbf{Navonil Majumder} is a PhD Student at Instituto Polit\'ecnico Nacional, Mexico. His research interests include natural language processing, machine learning, neural networks and deep learning. Contact him at \href{mailto:navo@nlp.cic.ipn.mx}{navo@nlp.cic.ipn.mx}.
\end{ieeecsAuthorBio}

\begin{ieeecsAuthorBio}
	\textbf{Soujanya Poria} is a presidential research fellow at Nanyang Technological University, Singapore. His research interests lie in sentiment analysis, multimodal interaction, natural language processing, and affective computing. Poria received a PhD in Computer Science and Mathematics from University of Stirling. Contact him at \href{mailto:sporia@ntu.edu.sg}{sporia@ntu.edu.sg}.
	
\end{ieeecsAuthorBio}

\begin{ieeecsAuthorBio}
	\textbf{Haiyun Peng} is a PhD Student at Nanyang Technological University, Singapore. His research interests include multilingual sentiment analysis, text representation learning, dialogue system and deep learning. Contact him at \href{mailto:peng0065@ntu.edu.sg}{peng0065@ntu.edu.sg}.
\end{ieeecsAuthorBio}

\begin{ieeecsAuthorBio}
	\textbf{Niyati Chhaya} is a senior research scientist at Adobe Research, India. Her research interests are in natural language processing and machine learning with a current research focus on affective content analysis. Chhaya completed her PhD from the University of Maryland. Contact her at \href{mailto:nchhaya@adobe.com}{nchhaya@adobe.com}.
\end{ieeecsAuthorBio}

\begin{ieeecsAuthorBio}
	\textbf{Erik Cambria} is an associate professor at Nanyang Technological University, Singapore. His research interests include natural language understanding, commonsense reasoning, sentiment analysis, and multimodal interaction. Contact him at \href{mailto:cambria@ntu.edu.sg}{cambria@ntu.edu.sg}.
\end{ieeecsAuthorBio}

\begin{ieeecsAuthorBio}
	\textbf{Alexander Gelbukh} is a professor at Instituto Politécnico Nacional, Mexico. His research interests lie in artificial intelligence, computational linguistics, sentiment and emotion analysis, computational morphology etc. Gelbukh received a PhD in Computer Science from All-Russian Institute for Scientific and Technical Information (VINITI). Contact him at \href{mailto:gelbukh@cic.ipn.mx}{gelbukh@cic.ipn.mx}.
\end{ieeecsAuthorBio}

\end{document}